\pdfoutput=1
\documentclass[letterpaper, 10 pt, conference]{ieeeconf}

\IEEEoverridecommandlockouts                              

\makeatletter
\let\NAT@parse\undefined
\makeatother

\usepackage{textcomp}
\usepackage{xcolor}
\usepackage{pifont}
\usepackage[per-mode = symbol]{siunitx}
\def\BibTeX{{\rm B\kern-.05em{\sc i\kern-.025em b}\kern-.08em
    T\kern-.1667em\lower.7ex\hbox{E}\kern-.125emX}}

\usepackage{amssymb,amsmath}

\usepackage[sort&compress,numbers,sectionbib]{natbib}

\usepackage{cite} 
\usepackage{booktabs}
\usepackage[font=small]{caption}
\usepackage{subcaption}
\usepackage{float}
\usepackage{graphicx}

\usepackage{verbatimbox}
\usepackage{multirow}
\usepackage{balance}
\usepackage{algorithm}
\usepackage{algorithmic}
\usepackage{comment}
\usepackage{tabularx}
\newcolumntype{Y}{>{\centering\arraybackslash}X} 

\usepackage{bbm}
\usepackage{float}
\usepackage{stfloats} 

\newcommand{\cmark}{\ding{51}}%

\usepackage{dsfont}

\newcommand{\figref}[1]{Fig.~\ref{#1}}

\usepackage[hyphens]{url}

\usepackage{hyperref}
\hypersetup{bookmarksopen,bookmarksnumbered,
pdfpagemode=UseOutlines,
colorlinks=true,
linkcolor=teal,
anchorcolor=teal,
citecolor=teal,
filecolor=teal,
menucolor=teal,
urlcolor=teal,
breaklinks=true
}
\usepackage[dvipsnames]{xcolor}
\definecolor{darkblue}{RGB}{0.15,0.15,0.55}
\definecolor{lightgrey}{RGB}{0.75,0.75,0.75}

\definecolor{myred}{RGB}{215,48,39}
\definecolor{myblue}{RGB}{69,117,180}
\definecolor{myorange}{RGB}{252,141,89}
\definecolor{mylightblue}{RGB}{145,191,219}

\definecolor{MYlightblue}{RGB}{217,95,2} 
\definecolor{MYdarkblue}{RGB}{117,112,179} 
\definecolor{MYgreen}{RGB}{27,158,119}

\usepackage{color,soul}
\usepackage{svg}

\newcommand{\xxnote}[3]{}
\ifx\hidenotes\undefined
  \usepackage{color}
  \renewcommand{\xxnote}[3]{\color{#2}{#1: #3}}
\fi

\usepackage[normalem]{ulem}
\useunder{\uline}{\ul}{}

\usepackage{bm}


\title{\LARGE \bf
Bi$^3$: A Biplatform, Bicultural, Biperson Dataset\\ for Social Robot Navigation$^*$}

\author{Andrew Stratton$^{1}$, Phani Teja Singamaneni$^{2, 3}$, Pranav Goyal$^{1}$, Rachid Alami$^{2}$, and Christoforos Mavrogiannis$^{1}$%
\thanks{$^{1}$Andrew Stratton, Pranav Goyal, and Christoforos Mavrogiannis are with the Department of Robotics, University of Michigan, Ann Arbor, USA.
         {\tt\footnotesize \{arstr, prgoyal, cmavro\}@umich.edu.}}%
\thanks{$^{2} $Phani Teja Singamaneni and Rachid Alami are affiliated with LAAS-CNRS, University of Toulouse, Toulouse, FR.
         {\tt\footnotesize \{ptsingaman, rachid.alami\}@laas.fr.}}
\thanks{$^{3} $Phani Teja Singamaneni is also affiliated with INRIA, University of Lorraine, Nancy, FR.
         {\tt\footnotesize \{phani-teja.singamaneni@inria.fr\}}}%
\thanks{$^*$Our dataset is available at: \url{https://fluentrobotics.com/bi3dataset/}.}
\thanks{This work was partially supported by the Horizon Europe Framework Programme grant 101070596 - euROBIN.}
}

\newcommand{\insertfig}{
\setcounter{figure}{0} 
\includegraphics[width=\linewidth]{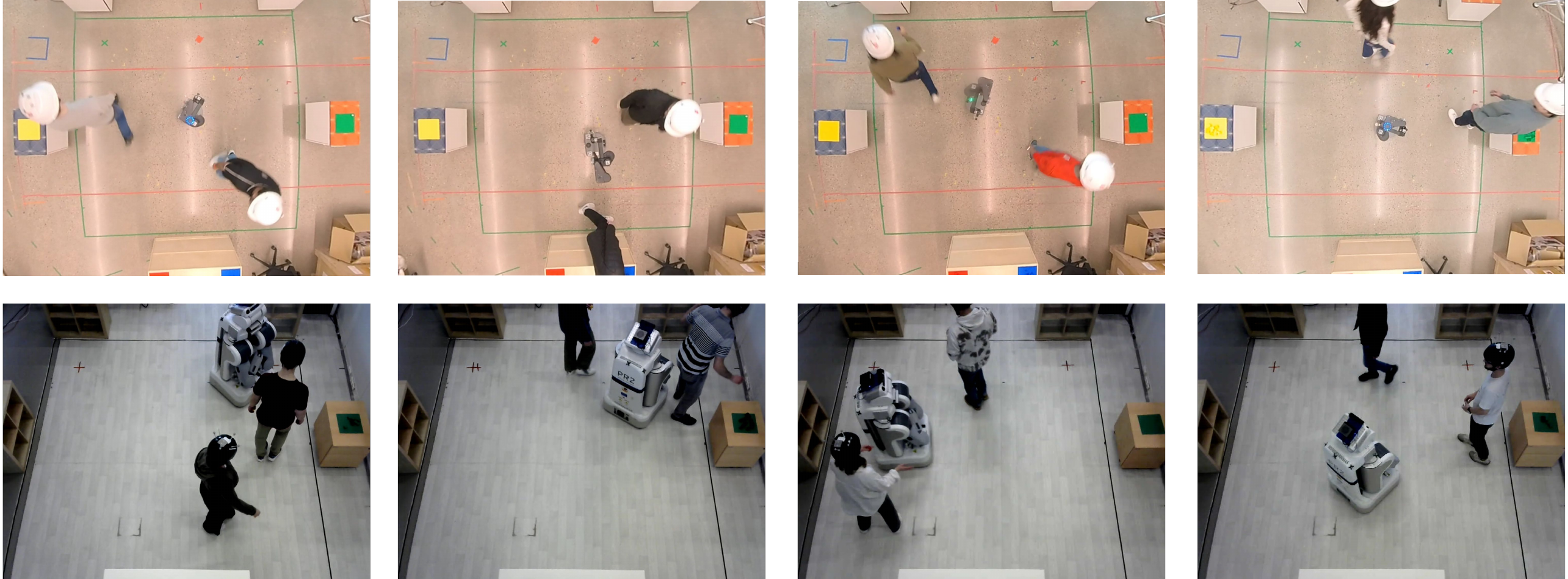}\captionof{figure}{We introduce Bi$^3$, a novel dataset which contains diverse navigation interactions in close-quarters settings captured with two robots at two sites with five fully autonomous robot controllers. The top row shows example user interactions with a Hello Robot Stretch at UM, whereas the bottom row shows interactions with a Willow Garage PR2 at LAAS-CNRS.}\label{fig:main_fig}}

\makeatletter
\apptocmd{\@maketitle}{\centering\insertfig}{}{}
\makeatother

\begin{document}

\maketitle

\thispagestyle{empty}
\pagestyle{empty}

\begin{abstract}
We contribute Bi$^3$, a dataset of social robot navigation among groups of people in a constrained lab space. Compared to prior data collection efforts for social robot navigation, our dataset is unique in that it features: an original experiment design giving rise to close navigation encounters between two humans and a robot; five different navigation algorithms; two different robot platforms; a diverse participant pool of 74 people recruited from two sites in the USA and France; multimodal data streams including 10.5 hours of human and robot ground-truth motion tracks, RGB video, and user impressions over robot performance. Our analysis of the collected dataset through metrics like interaction density and human velocity suggests that Bi$^3$ represents a benchmark of unique diversity and modeling complexity. Bi$^3$ contributes towards understanding how humans and robots can productively mesh their activities in constrained environments, and can be a resource for training models of human motion prediction and robot control policies for navigation in densely crowded spaces.

\end{abstract}

\section{Introduction}

A critical component of any social robot navigation (SRN) system is a model for how people move around a robot~\citep{core-challenges2021,singamaneni2023survey}. This has motivated extensive research on human motion prediction (HMP)~\citep{poddar2023crowd, gupta2018social, sophie19, trajectron, salzmann2023hst}, the problem of estimating future interactions among groups of people given observations of their behavior over a window of the past. Much of the recent work on HMP makes use of deep-learning techniques due to their ability to represent complex interactions in high-dimensional spaces. The success of these techniques relies on the closeness between the dataset they are trained on and the distribution of interactions they encounter during deployment. This has motivated extensive data collection efforts involving the deployment of cameras, robots, or sensorized devices in crowded human spaces \citep{PellegriniESG09, lerner_ucy, brscic_atc, alexandre_sdd, wang_tbd_2024, martin2021jrdb, yan_lcas}.

\begin{table*}[hbt!]
\centering
\caption{Comparison between our dataset and other HMP and SRN datasets. Blank columns are not applicable to that dataset row.}
\label{tab:literature}
\resizebox{\linewidth}{!}{%
\renewcommand{\arraystretch}{0.5}
\begin{tabular}{l| c c c c c c c c}
\toprule
\textbf{Dataset} & \textbf{Scene} & \textbf{Multiple Sites} & \textbf{Duration} & \textbf{Pose Tracking} & \textbf{Goals} & \textbf{Multiple Robot Behaviors} & \textbf{\# of Robot Platforms} & \textbf{User Impressions} \\
\midrule
ETH~\citep{PellegriniESG09} & Street & \cmark & 25' & Camera &  &  & 0 & \\
\midrule
UCY~\citep{lerner_ucy} & Street &  & 20' & Camera &  &  & 0 & \\
\midrule
Edinburgh~\citep{edinburgh_09} & Outdoor &  & 4 months & Camera &  &  & 0 & \\
\midrule
ATC~\citep{brscic_atc} & Indoor &  & 41 days & 3D range &  &  & 0 & \\
\midrule
SDD~\citep{alexandre_sdd} & Outdoor & \cmark & 5 days & Camera &  &  & 0 & \\
\midrule
TBD~\citep{wang_tbd_2024} & Indoor &  & 12.7 hrs & Camera &  &  & 1 & \\
\midrule
JRDB~\citep{martin2021jrdb} & Indoor/outdoor & \cmark & 64' & Camera &  & \cmark & 1 & \\
\midrule
L-CAS~\citep{yan_lcas} & Indoor & \cmark & 49' & 3D LiDAR &  & & 1 & \\
\midrule
SCAND~\citep{karnan2022scand} & Indoor/outdoor & \cmark & 8.7 hrs &  &  &  & 2 & \\
\midrule
MuSoHu~\citep{nguyen_toward_2023} & Indoor/outdoor & \cmark & 20 hrs &  &  &  & 0 & \\
\midrule
INTERACT~\citep{posetron} & Lab &  & 9 hrs & MoCap & \cmark &  & 1 & \\
\midrule
THOR~\citep{rudenko_thor_2020} & Lab &  & 1 hr & MoCap & \cmark &  & 1 & \\
\midrule
THOR MAGNI~\citep{schreiter_thor-magni_2025} & Lab &  & 3.5 hrs & MoCap & \cmark & \cmark & 1 & \\
\midrule
HuRON~\citep{hirose_sacson} & Office &  & 75 hrs &  &  & \cmark & 1 & \\
\midrule
CrowdBot \citep{paezgranados2022pedestrianrobot} & Outdoor &  & 1 hr &  &  & \cmark & 1 & \\
\midrule
\textbf{Bi$^3$} (Ours) & Lab & \cmark & 10.5 hrs & MoCap & \cmark & \cmark & 2 & \cmark \\
\bottomrule
\end{tabular}
}
\end{table*}

We observe a few critical gaps in existing datasets supporting SRN and HMP. First, the vast majority of datasets often feature very sparse and distant interactions among people~\citep{scoller2019cv, amirian2020opentraj}. To truly integrate in realistic human spaces, robots need to be prepared to respond adequately to close human-robot navigation encounters. Second, many datasets lack robot deployments \citep{nguyen_toward_2023, PellegriniESG09, lerner_ucy, brscic_atc, alexandre_sdd, yan_lcas}, and instead include data collected via overhead cameras. Robots deployed in the field, however, will need to leverage onboard sensing to perceive and react to people, leading to additional perception challenges \citep{martin2021jrdb, yan_lcas}. When robots are part of a data collection effort, they are often deployed ``in the wild.'' While it is important to train systems on real-world, unscripted interactions, the lack of control for external variables influencing human behavior around robots could lead to poor data associations performed during model training \citep{Fujimoto2018OffPolicyDR}. Many robot-deployment datasets are collected via remote teleoperation of a robot, without any autonomy involved \citep{martin2021jrdb, karnan2022scand, schreiter_thor-magni_2025}. This does not allow for the extraction of insights about how algorithm embodiment would impact human-robot interaction. Many datasets recording human-robot navigation encounters lack any information about the impressions of people around them. This impedes the extraction of insights about how to assess and design \emph{socially competent} robot behavior, which is ultimately the main goal of SRN~\citep{core-challenges2021}. Finally, the perceived social competence of robot behavior is often a factor of robot morphology and cultural norms present in a deployment region \citep{walters_design}. Thus, it is important to enable researchers to introduce these dependencies into the model training and validation.

Motivated by these gaps (see table \ref{tab:literature} for a summary), we introduce Bi$^3$, a new SRN dataset collected during our IRB-approved user study~\citep{stratton2026humanmotionpredictionquality}, featuring: two different platforms available to numerous labs around the world (Hello Robot Stretch, Willow Garage PR2, see fig. \ref{fig:main_fig}); 10.5 hours of natural, close navigation encounters between a robot and 74 participants, recruited from two sites in distinct regions of the world (University of Michigan in Ann Arbor, USA and LAAS-CNRS in Toulouse, France); a wide range of autonomous robot behavior resulting from the embodiment of five different HMP models on a model predictive control (MPC) scheme; high-accuracy human and robot motion tracks; top-down RGB video and egocentric depth images; prediction-control pairs for every control cycle; user impressions about the robot collected via Likert-style questionnaires and open-ended comments. Our analysis shows that Bi$^3$ has very dense, high-velocity human interactions, diverse human and robot motion, and diverse human preferences.

\section{Related Work}

Initial dataset efforts for analyzing human navigation interactions, including the smaller-scale ETH \citep{PellegriniESG09} and UCY \citep{lerner_ucy} datasets, as well as the larger-scale Edinburgh \citep{edinburgh_09}, ATC \citep{brscic_atc} and Stanford Drone (SDD) \citep{alexandre_sdd} datasets, focused mainly on open, public spaces. While these datasets do not include robots in their scenes, and are thus less suited to studying human-robot navigation interactions, they have been used extensively for the training and benchmarking of HMP algorithms \citep{gupta2018social, sophie19, trajectron, salzmann2023hst, amirian2020opentraj} and multi-target tracking \citep{Sadeghian2017TrackingTU, Fuchs2019EndtoendRM}.

Several datasets have been collected in similar spaces using robots or sensorized devices. The TBD dataset \citep{wang_tbd_2024} contributes to the data collection pipeline with a portable data collection and automated labeling system. The JRDB \citep{martin2021jrdb} and L-CAS \citep{yan_lcas} datasets collected sensor data aboard a mobile robot to enable studying robot-centric perception challenges, including tracking \citep{vendrow2023jrdb, yan_lcas}, activity recognition \citep{ehsanpour2022jrdb}, as well as HMP \citep{saadatnejad2023jrdb, salzmann2023hst}. The SCAND \citep{karnan2022scand} dataset leveraged teleoperation with two mobile robots to collect expert SRN demonstrations for training imitation learning policies, while the MuSoHu \citep{nguyen_toward_2023} dataset has human demonstrators equipped with sensors. These datasets, owing to their collection in public areas with ample space, are sparse in regard to close-quarters human-robot interactions \citep{amirian2020opentraj}. Bi$^3$ is collected in a constrained workspace with structured human and robot tasks that forced consistent close encounters, making it more effective for learning models for deployment in many indoor settings where space is limited.

The datasets most similar to ours are those which are collected through experiments in lab settings, allowing for finer control over the interactions occuring in the data. The INTERACT \citep{posetron} dataset has three humans interacting with a robot, but focuses on human skeleton trajectory prediction during collaborative assembly tasks. Our work instead uses assembly tasks to motivate challenging multiagent navigation interactions. The THOR \citep{rudenko_thor_2020}  and THOR MAGNI \citep{schreiter_thor-magni_2025} datasets also focus on multiagent interactions resulting from humans and a robot moving between stations to complete tasks in a shared workspace. They prioritize high task diversity, larger numbers of humans, and many sensor modalities in support of HMP with additional human features \citep{zheng_gimo, li_fusion, rudenko2019-predSurvey}. Our dataset instead adds diversity of close interactions by recording human-robot navigation encounters resulting from five different algorithms, in two locations, and with two robot platforms, factors which increase the applicability of our data to a wider range of downstream robot deployments. Crucially, our dataset also includes human-centered attributes through the recording of user impressions of robot sociability \citep{carpinella} and workload \citep{hart1988development} which can enable the development and validation of human-centered algorithms.

\section{Social Robot Navigation System}\label{system}

\begin{figure*}[t]
    \centering
    \includegraphics[width=0.9\linewidth]{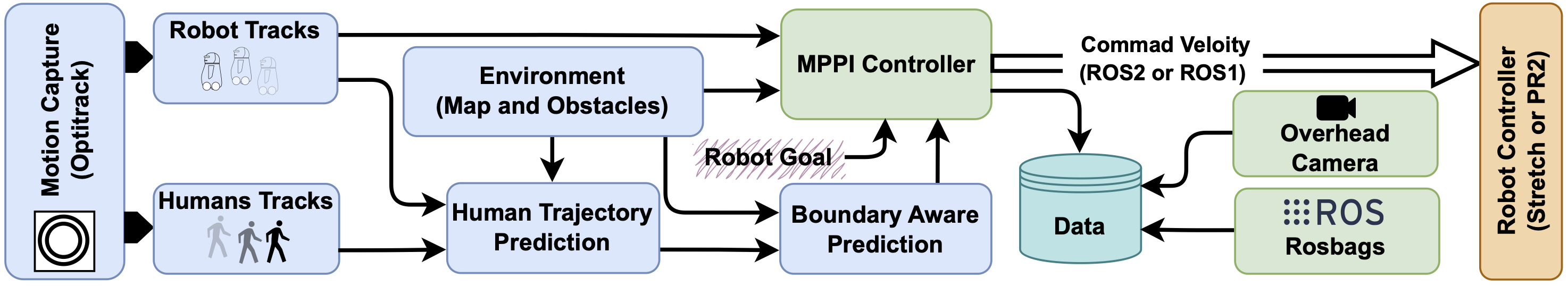}
    \caption{Architecture and data collection pipeline.}
    \label{fig:pipeline}
\end{figure*}

Our SRN system is implemented as an MPC architecture due to its flexibility, modularity, and community adoption~\citep{poddar2023crowd, stratton2025complexity, samavi2023sicnav, dr-mpc-deep-residual}. 
We formalize our domain, and discuss the HMP models used in our data collection. A diagram of our system and data collection pipeline is in figure \ref{fig:pipeline}.

\subsection{Domain} 

We consider a robot navigating among $n \geq 1$ humans in a workspace $\mathcal{W}\subseteq \mathbb{R}^2$. We describe the state of the robot as $s^r\in\mathcal{W}$ and the state of human $i\in \{1,\dots, n\}$ as $s^i$. The state of the robot evolves according to dynamics $s^r_{t+1}=g(s^r_t,u_t)$, where $u_t$ is a control action (speed, and steering angle), drawn from a space of controls $\mathcal{U}$. The robot starts from an initial configuration $s^r_0$ and moves towards a goal $d^r$ by following a policy $\pi^r$, while humans navigate from their initial configurations $s^i_0$ towards their goals, $d^i$ by following a policy $\pi^i$, $i\in\mathcal{N}$; agents' goals and policies are unknown to one another.  
We assume that the robot perfectly observes human states at all times. We focus on the problem of designing a robot policy $\pi^r$, such that the robot reaches its goal $d^r$ in a safe, efficient, and socially compliant fashion.

\subsection{Model Predictive Control for Social Robot Navigation} 

Our MPC is an implementation of the robot policy $\pi^r$ as a receding-horizon control optimization over a horizon $T$:
\begin{equation}
\begin{split}
    u^*_{t:T} & = \arg\min_{u_{t}\in\mathcal{U}} \sum_{t}^{T-1}\mathcal{J}(s^r_{t+1}, s^{1:n}_{t+1}) \\
    & s.t.\  s^r_{t+1} = g(s^r_t, u_t) \\
    & \qquad s^{1:n}_{t+1} = f(s^r_{t-h:t}, s^{1:n}_{t-h:t})\mbox{,} \label{eq1}
\end{split}
\end{equation}
where $\mathcal{J}$ is a cost function capturing qualities of SRN, and $f$ is an HMP model that takes as input the state history of the robot $s^r_{t-h:t}$ and humans $s^{1:n}_{t-h:t}$ over a window $h$, and outputs a prediction of human motion, $s^{1:n}_{t+1}$. Following prior work~\citep{mavrogiannis2023winding, poddar2023crowd}, we model cost $\mathcal{J}$ as:
\begin{equation} \label{eq2}
    \mathcal{J}(s^r, s^{1:n}) = a_g \mathcal{J}_g(s^r) + a_d \mathcal{J}_d (s^r, s^{1:n}) + a_o \mathcal{J}_o (s^r)\mbox{,}
\end{equation}
where $\mathcal{J}_g (s^r) = (s^r - d^r)^2$ penalizes distance to the robot's goal, $\mathcal{J}_d(s^r, s^{1:n}) = \sum_{i=1}^{n} A_d^2 (s^r, s^i)$, penalizes violations to the personal space as done in prior work~\citep{kirby_thesis}, and $\mathcal{J}_o (s^r) = \mathds{1}(A_R(s^r) \cap \mathcal{W}^{c} \neq \emptyset)$ penalizes violations of the workspace boundary, where $A^r(s^r)$ is the volume occupied by the robot at pose $s^r$. Weights $a_g$, $a_d$, and $a_o$ represent the relative importance of each cost. 

\subsection{Human Modeling} 
\begin{figure*}[t]
  \centering
  \newlength{\sqw}
  \setlength{\sqw}{\dimexpr\textwidth/6\relax}

  \setlength{\tabcolsep}{0pt} 
  \begin{tabular}{@{}cccccc@{}}
    \subcaptionbox{NP\label{fig:blind_path}}{%
      \includegraphics[width=\sqw]{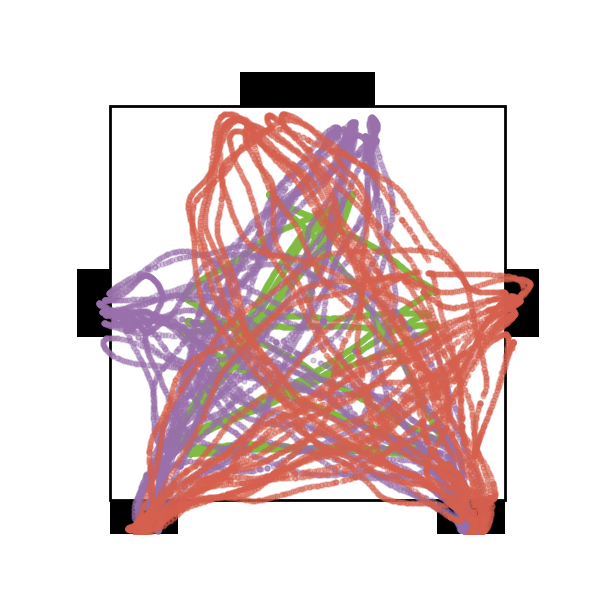}} &
    \subcaptionbox{ST\label{fig:static_path}}{%
      \includegraphics[width=\sqw]{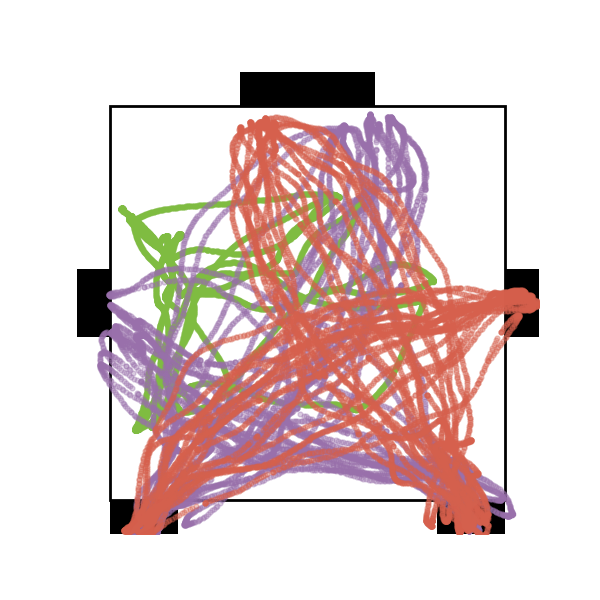}} &
    \subcaptionbox{CV\label{fig:cv_path}}{%
      \includegraphics[width=\sqw]{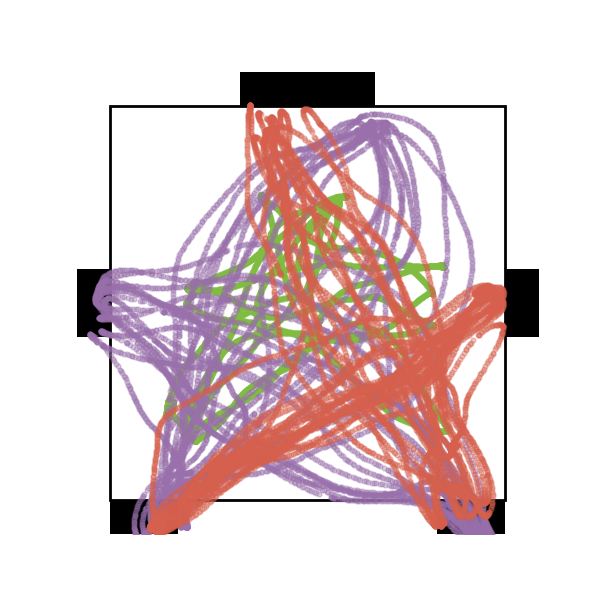}} &
    \subcaptionbox{CoHAN\label{fig:cohan_path}}{%
      \includegraphics[width=\sqw]{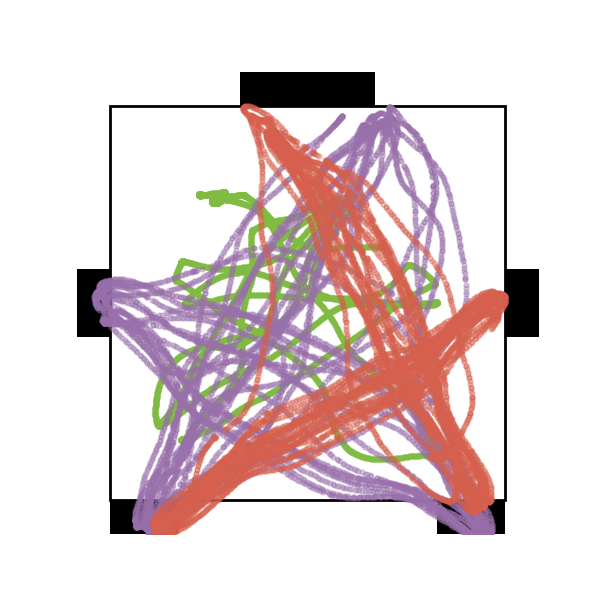}} &
    \subcaptionbox{HST\label{fig:hst_path}}{%
      \includegraphics[width=\sqw]{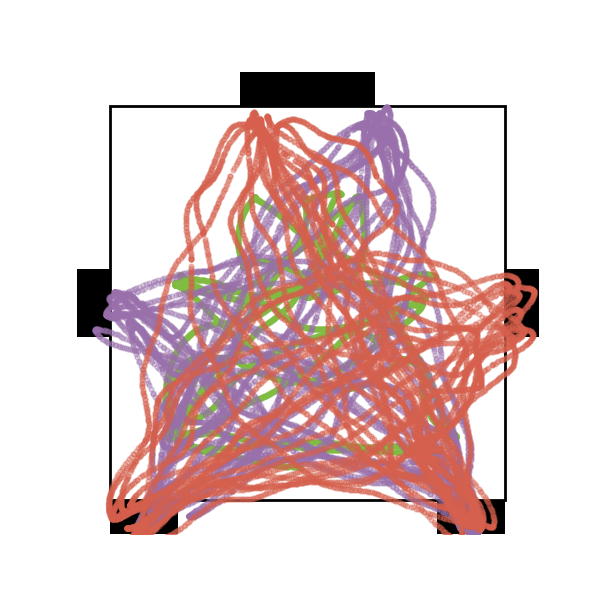}} &
    \subcaptionbox{Key\label{fig:key_path}}{%
      \includegraphics[width=0.5\sqw]{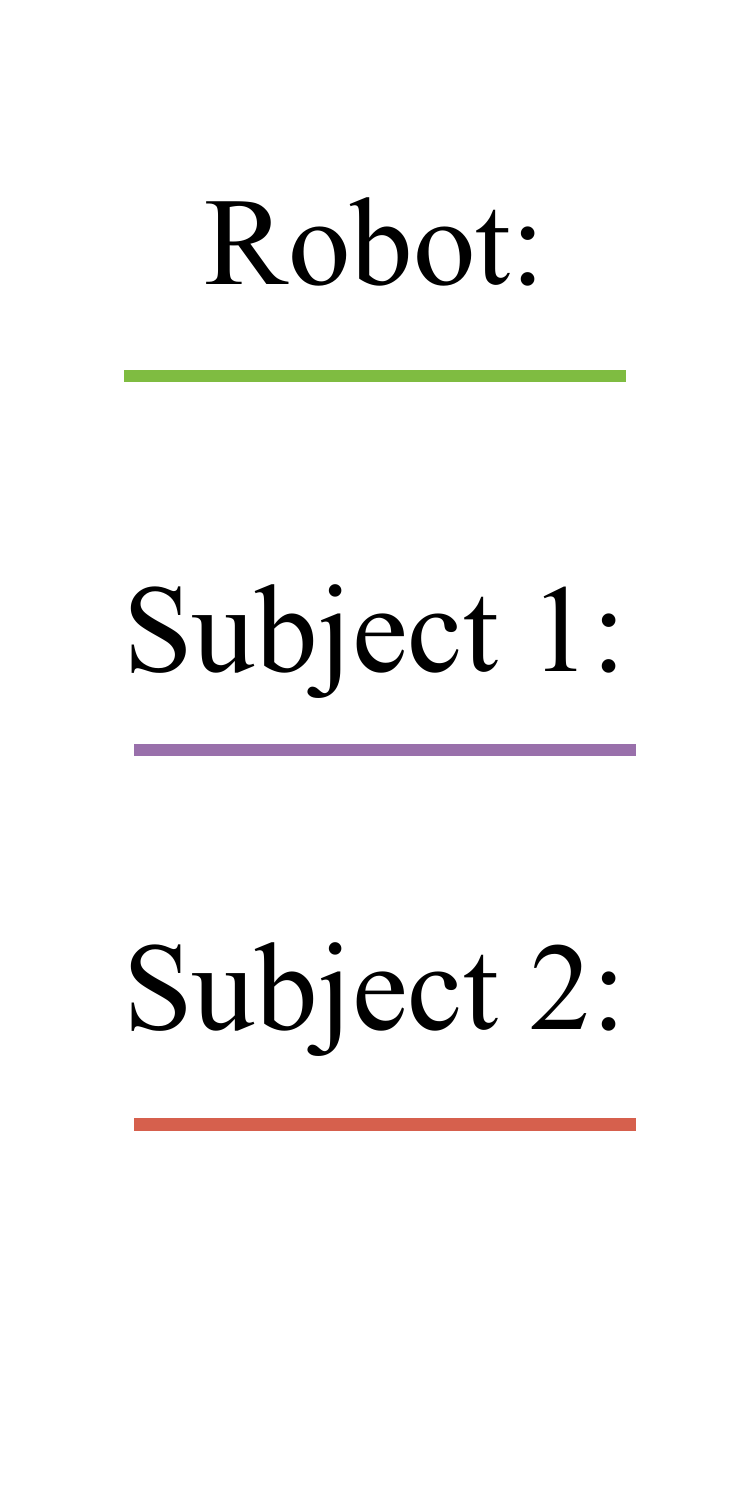}}
  \end{tabular}

  \caption{Human and robot paths over a single trial for all five controllers. Despite the humans and robot having the same sequence of workspace goals for all five, the resulting paths are substantially different.}
  \label{fig:paths}
\end{figure*}

Our system's modular design allows the integration of any HMP model. To capture a wide diversity of robot human-robot navigation encounters (examples are shown in fig. \ref{fig:paths}), we integrated five different HMP models into our MPC:

\textbf{No Prediction (NP)}: The robot does not perceive the humans navigating around it. NP is representative of the common practice in industrial robotics of having unreactive robots \citep{HANNA2022102386}.

\textbf{Static (ST):} Humans are treated as static obstacles, as in many costmap-based SRN approaches \citep{cohan, best_of_both}. In close-quarters scenarios in which agents rely on each other moving out of their paths to make progress, ST represents a highly conservative model.

\textbf{Constant Velocity (CV)}: Humans are treated as dynamic obstacles with no human-aware modeling. CV has repeatedly been shown to be an effective model of human navigation not only in terms of prediction accuracy \citep{scoller2019cv}, but also for downstream use in SRN \citep{poddar2023crowd, stratton2025complexity, cohan, social_costmap}.

\textbf{Human Scene Transformer (HST)}: HST \citep{salzmann2023hst} is a transformer-based generative model which represents the large class of deep learning-based prediction methods \citep{salzmann2023hst, trajectron, sophie19, gupta2018social}. We choose the HST for its strong performance on established benchmarks \citep{PellegriniESG09, lerner_ucy, martin2021jrdb}.

\textbf{CoHAN}: CoHAN \citep{cohan, hateb2, singamaneni2022watch} is a trajectory planner that plans robot and predicted human trajectories jointly, representing the paradigm of \emph{cooperative collision avoidance}~\citep{vandenberg, sun_brne, cohan, hateb2, singamaneni2022watch,trautmanijrr,mavrogiannis_social_2022}. It optimizes the trajectories using social costs including human-human and human-robot safety, human visibility and speed, and obstacle avoidance.

\subsection{Implementation Details}

\begin{figure}
    \centering
    \includegraphics[width=0.9\linewidth]{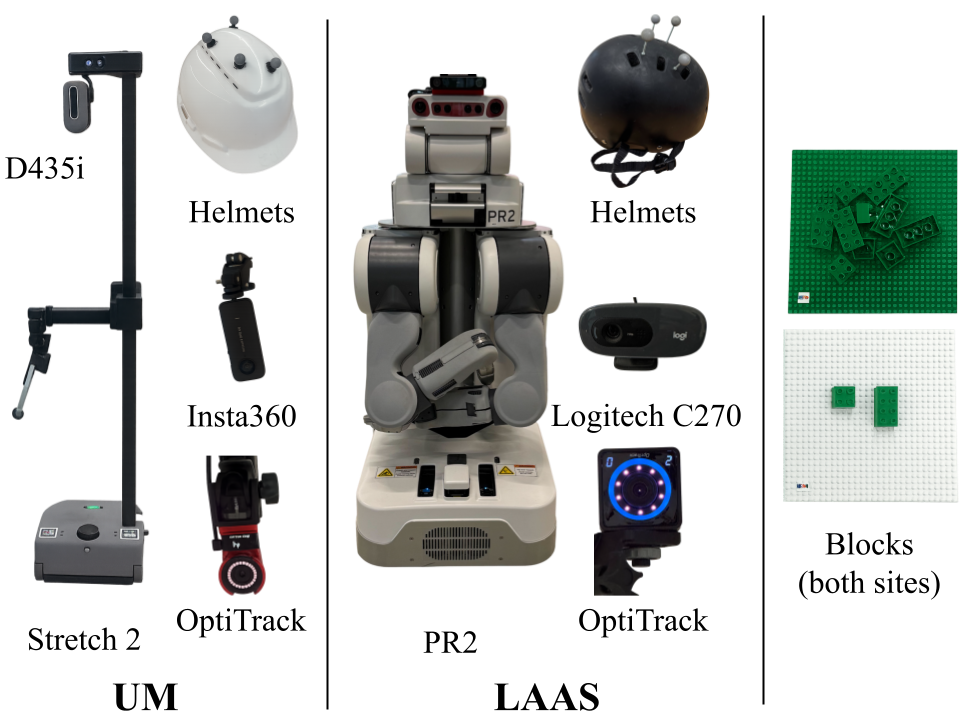}
    \caption{Hardware used for the data collection.}
    \label{fig:apparatus}
\end{figure}

\textbf{Hardware}: We instantiated our MPC under all HMP models on two different robot platforms: a Hello Robot Stretch at UM, and a Willow Garage PR2 at LAAS. The two platforms are opposite in their size ($330 \times 340 \times 1410mm$ v.s. $668 \times 668 \times 1330mm$) and mass (24.5 kg v.s. 226.8 kg), which we anticipated would affect human perceptions and navigation interactions. Overhead motion capture systems (OptiTrack at both sites) provided high-accuracy robot localization and people tracking at 120 Hz. The coverage spanned the entire workspace at both sites; thus, the commonly reported \emph{Tracking Duration} metric \citep{rudenko_thor_2020, schreiter_thor-magni_2025, wang_tbd_2024, PellegriniESG09, lerner_ucy, brscic_atc} is simply the average length of our trials, which are $183.67s$ at UM and $226.22s$ at LAAS. HMP models and MPC were verified and tuned on hardware at each site to ensure consistent behavior. Fig.~\ref{fig:apparatus} shows the equipment used at both sites to collect the data.

\textbf{HMP}: All models used an $h=8$ timestep history and a prediction horizon of $T=12$, with a timestep of $0.4s$, matching standard evaluation settings in the HMP literature \citep{gupta2018social, sophie19, trajectron, salzmann2023hst}. Predictions violating the workspace boundary were truncated and replaced with the latest valid within-boundary predictions. The HST checkpoint was trained on the ZARA1 portion of the ETH/UCY dataset, producing $m=20$ prediction modes, the most likely of which was used. To match other models, CoHAN's backoff behavior and FoV restriction on planning were removed.

\textbf{MPC}: Our control architecture is based on a GPU-parallelized implementation of Model Predictive Path Integral (MPPI) control~\citep{mppi}, using 2000 trajectory samples. The maximum speed of the robot was capped to $0.3$\si{\metre\per\second}, as it was found to be a safe limit for our setup. The controller uses a differential-drive kinematic model with a control space consisting of linear and angular velocities at both sites. The full prediction and control loop closed at different frequencies depending on the computational burden of the HMP model, with NP, ST, and CV operating at 50 Hz, HST at 20 Hz, and CoHAN at 8 Hz.

\textbf{Computation and communications}: All computation was performed on lab computers, with velocity commands and sensor data transmitted wirelessly. All communication was done using ROS (ROS2 for Stretch and ROS1 for PR2) with the MPPI running in ROS2 and CoHAN in ROS1. This required ROS bridges to convert messages between the two ROS versions for the two robot platforms.

\textbf{Software}: The system code is modularized so prediction component can be easily swapped with any trajectory prediction model operating on positions. Furthermore, the controller implementation is not platform specific: it works with any low-level controller which executes velocity commands.
\section{Data Collection}\label{collection}

We conducted an ethics approved user study (University of Michigan IRB HUM00259961; Universit\'e Toulouse Ethics Approval 2024\_917) to collect our data through controlled experiments. For the full details and in-depth analysis of human-robot interactions and user impressions, we refer the reader to our user-study paper~\citep{stratton2026humanmotionpredictionquality}. Here, we summarize our scenario design and detail our data collection procedure.

\begin{figure}
    \centering
    \includegraphics[width=\linewidth]{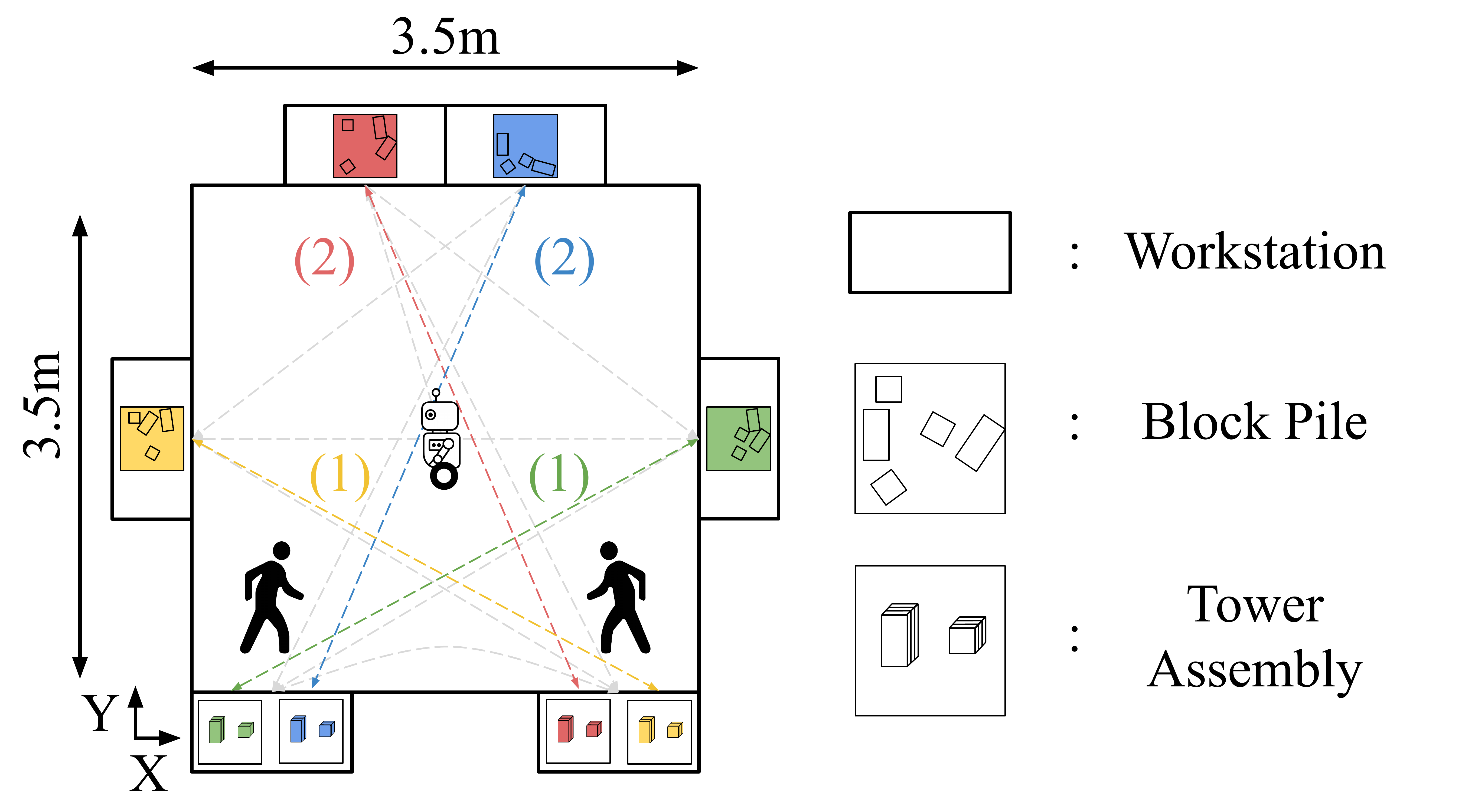}
    \caption{Experimental setup. Subjects begin and end in front of the assembly stations while the robot begins in the center of the workspace. The blocks labeled (1) are assembled by subjects followed by those labeled (2), while the robot traverses between the goals along the path in gray.}
    \label{fig:setup}
\end{figure}

\subsection{Scenario Design}

Collecting data on group interactions with a navigating robot in a lab environment is challenging: compared to data collections in the field, more care must be taken to ensure natural human motion and diverse interactions even during highly structured scenarios. To that end, we build upon prior work instantiating industrial teamwork settings in a lab environment~\citep{landolfi_working_2023, mavrogiannis_social_2022}.

Our scenario involves two subjects repeatedly traversing an open workspace between workstations along the perimeter, picking blocks from piles, traversing, and placing them one-by-one to assemble towers. To match many industrial and home scenarios which necessitate close-quarters interactions, we made the traversable portion of the workspace $3.5 \times 3.5m$. To limit subject interactions to navigation, participants were each responsible for their own set of two out of four total towers and instructed not to talk. These sets, along with a specified ordering for completing the towers in each set, were designed to maximize interaction density by ensuring participants consistently crossed paths with each other and the robot as shown in Fig. \ref{fig:setup}.

While subjects assembled towers, the robot navigated through a randomly generated sequence of workstations. To instigate regular crossings with both subjects, the sequence required the robot to cross the y-axis of the workspace between each goal (Fig. \ref{fig:setup}). The sequence was flipped across the y-axis each trial to reduce learning effects and further increase the diversity of interactions. To give participants the impression the robot also completed tasks even though it did not physically interact with the environment, they were told the robot was monitoring their progress.

\subsection{Collection Procedure}

In each experiment, two subjects completed five trials, each with a distinct HMP model. Subjects began by reading and signing a consent form, after which they were briefed on the scenario and completed a practice trial in which the robot remained static at its starting position. They then completed the five trials, with impressions being collected via online surveys in between each trial. To account for ordering effects, the condition order was determined using a balanced Latin square. Upon completing all five trials and surveys, subjects completed demographic surveys, were given compensation or gifts, and debriefed with an information sheet describing the specific details of the experiment.

Subjects started and ended each trial at `X' shapes marked on the ground in front of the two assembly stations, while the robot began in the center of the workspace facing towards the subjects. The task began once verbal confirmation of readiness was received from both subjects, and the robot began moving. It was considered complete when both subjects completed their towers and returned to their starting positions. Each experiment took approximately 50 minutes and yielded approximately 17 minutes of data.

\section{The Bi$^3$ Dataset}\label{data}

The dataset consists of 37 experiments (17 taking place at UM and 20 at LAAS) involving a total of 74 subjects. Sample paths from the dataset are shown in~\figref{fig:paths}. Bi$^3$ can be downloaded at: \url{https://fluentrobotics.com/bi3dataset/}.

\subsection{Objective Data}

Our dataset includes raw sensor data including robot and human transforms from motion-capture as well as video recorded from cameras at both sites (Insta360 at UM and Logitech C270 at LAAS). Both cameras faced top-down, with the UM camera in the center of the workspace and the LAAS camera at one side. Video was synchronized with saved ROS bags (ROS2 in UM and ROS1 in LAAS) using recording metadata and message timestamps. Data from UM also includes depth camera data from the Stretch's onboard Intel Realsense D435i, which is mounted at the top of the robot. Top-down camera data was captured at 30 Hz at both sites, while the egocentric camera published at 15 Hz. In addition to raw sensor data, we captured additional information calculated as part of the control code, including HMP model outputs, the robot goal, and robot velocity commands. These were stored in JSON files, saved directly at each control loop during the experiment. Finally, we contribute auxillary information regarding the locations of the workspace boundary, workstations, human goals, and human tasks, which have been shown to improve prediction performance in similar domains \citep{schreiter_thor-magni_2025}.

\subsection{User Impressions}

We move beyond prior SRN datasets by also including human impressions of the robot performance per HMP model. Human impressions have emerged as not only critical for human-robot interaction studies, but also as a modality for training data-driven models for interactive tasks \citep{christiano_rlhf, wang2022feedbackefficient}. Our surveys included quantitative impressions, collected via the Robotic Social Attributes Scale (RoSAS) \citep{carpinella}, and the NASA Task Load Index (TLX) \citep{hart1988development}. The full RoSAS survey consists of 18 Likert-style questions which rate the amount an attribute applied to the robot. Six questions capture \emph{Discomfort}, six capture \emph{Competence}, and six capture \emph{Warmth}; we omitted the Warmth questions because they do not readily apply to SRN interactions, and used a nine-point scale. The TLX consists of 21-point Likert-style questions in which the user must rate their perceived workload along six different axes; we used all six questions. Our surveys also collected qualitative impressions via short open-ended questions requesting optional additional thoughts. These have been shown to enable the extraction of critical insights about user experience that are not well captured by standard scales~\citep{mavrogiannis_social_2022}. The dataset contains 285 survey responses and 78 comments from 57 participants (17 at UM, 40 at LAAS), as participants at UM were able to opt out of including their survey responses.
\section{Analysis} \label{analysis}

We discuss the quality and diversity of the human motion, robot motion, and human impressions of Bi$^3$.

\begin{figure}[t]
  \centering
  \includegraphics[width=\columnwidth]{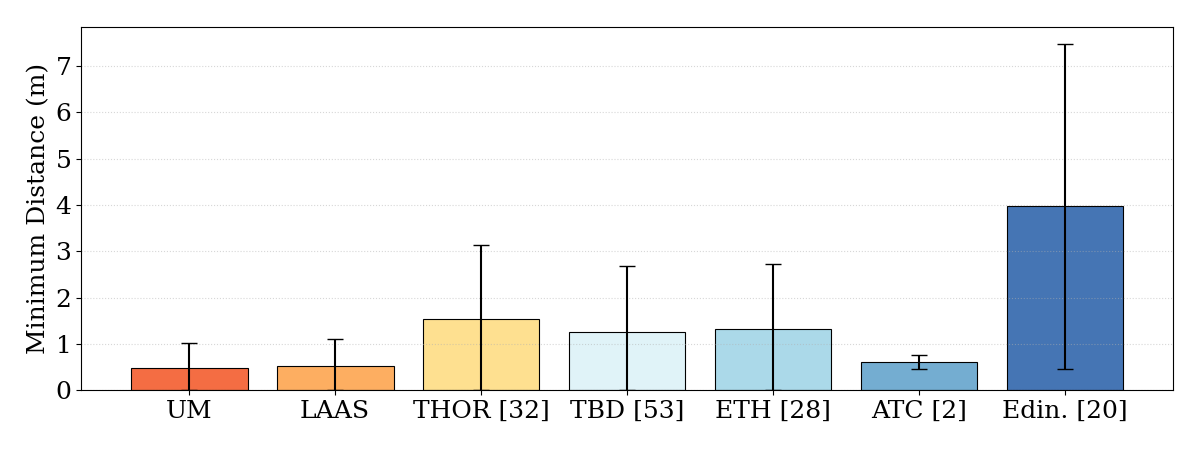}
  \includegraphics[width=\columnwidth]{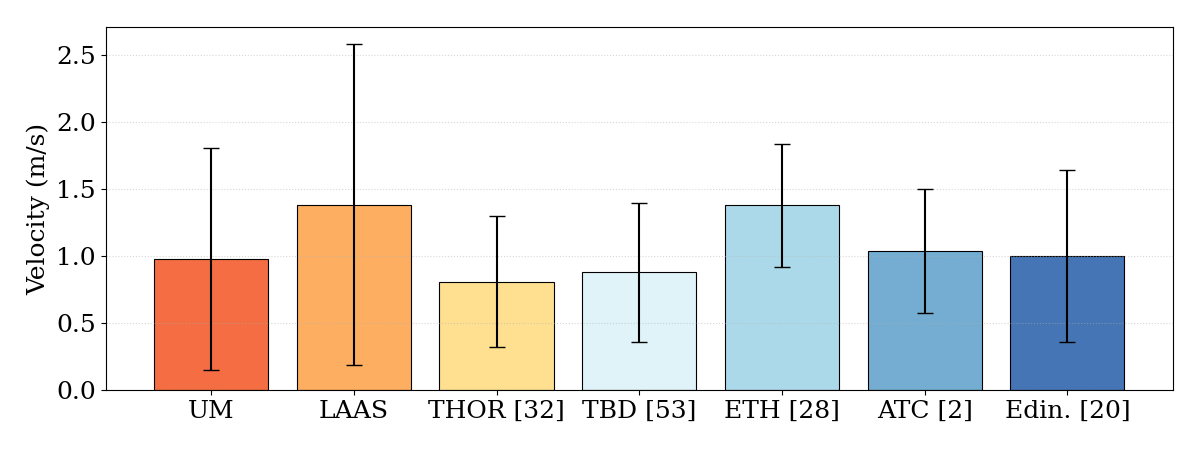}
  \caption{Human motion statistics between our dataset and several others (bars indicate means and errors are standard deviations).}
  \label{fig:hmp_stats}
\end{figure}

\textbf{Human Motion Profile}. Our data collection, through its large bicultural subject pool, diverse robot behaviors, and distinct robot embodiments, generated a highly diverse set of human navigation behaviors. We quantify this through the following metrics proposed in prior work \citep{rudenko_thor_2020}:
\begin{itemize}
    \item \textit{Minimum Distance} ($m$): The minimum distance between the two humans over a single tracklet, which is commonly used in SRN work related to personal space, or \emph{proxemics} \citep{gao-eval, stratton2025complexity, mavrogiannis_social_2022}. Lower minimum distance indicates closer interactions.
    \item \textit{Velocity} ($m/s$): The velocity of each human. Lower velocities allow more time for other agents to adapt their trajectories cooperatively, allowing for potentially smoother interactions.
\end{itemize}
We compare against related datasets in figure \ref{fig:hmp_stats}. We verify the high diversity of human motion in Bi$^3$ through its substantially higher standard deviation of recorded human velocity. Additionally, our data has the lowest minimum distance at both sites and the highest velocity at LAAS. These attributes make human trajectories difficult to predict, demonstrating the challenge Bi$^3$ presents.

\textbf{Robot Motion Profile}. We quantify the differences in robot motion arising from distinct HMP models using the following metrics common to the SRN literature \citep{gao-eval, core-challenges2021, mavrogiannis_social_2022}:
\begin{itemize}
    \item \textit{Minimum Distance} ($m$): Higher Minimum Distance is often connected to discomfort or unsafe behavior in prior work \citep{gao-eval}.
    \item \textit{Acceleration} ($m/s^2$): Higher average acceleration is indicative of more sudden movements, which have the potential to disrupt or startle co-navigating humans.
    \item \textit{Velocity} ($m/s$): Higher velocities indicate the robot moved efficiently, with less time slowed for humans.
    \item \textit{Path Efficiency} (ratio): The ratio of the distance between the trajectory start and endpoints and the cumulative distance traveled. Higher Path Efficiency can indicate less cooperative robot motion, as the robot does not adjust its paths for humans.
\end{itemize}

\begin{figure}[t]
  \centering
  \includegraphics[width=0.48\columnwidth]{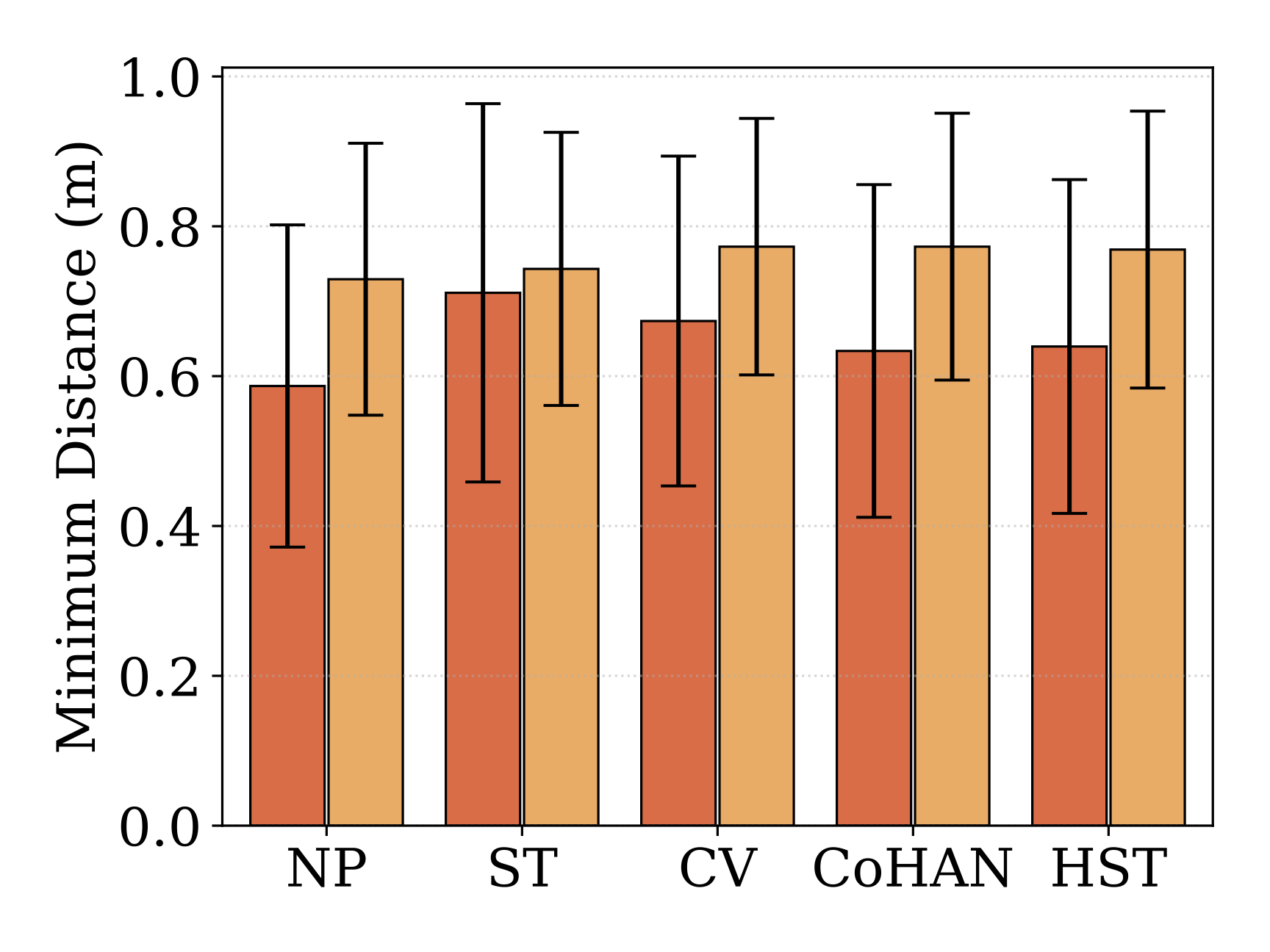}%
  \hfill
  \includegraphics[width=0.48\columnwidth]{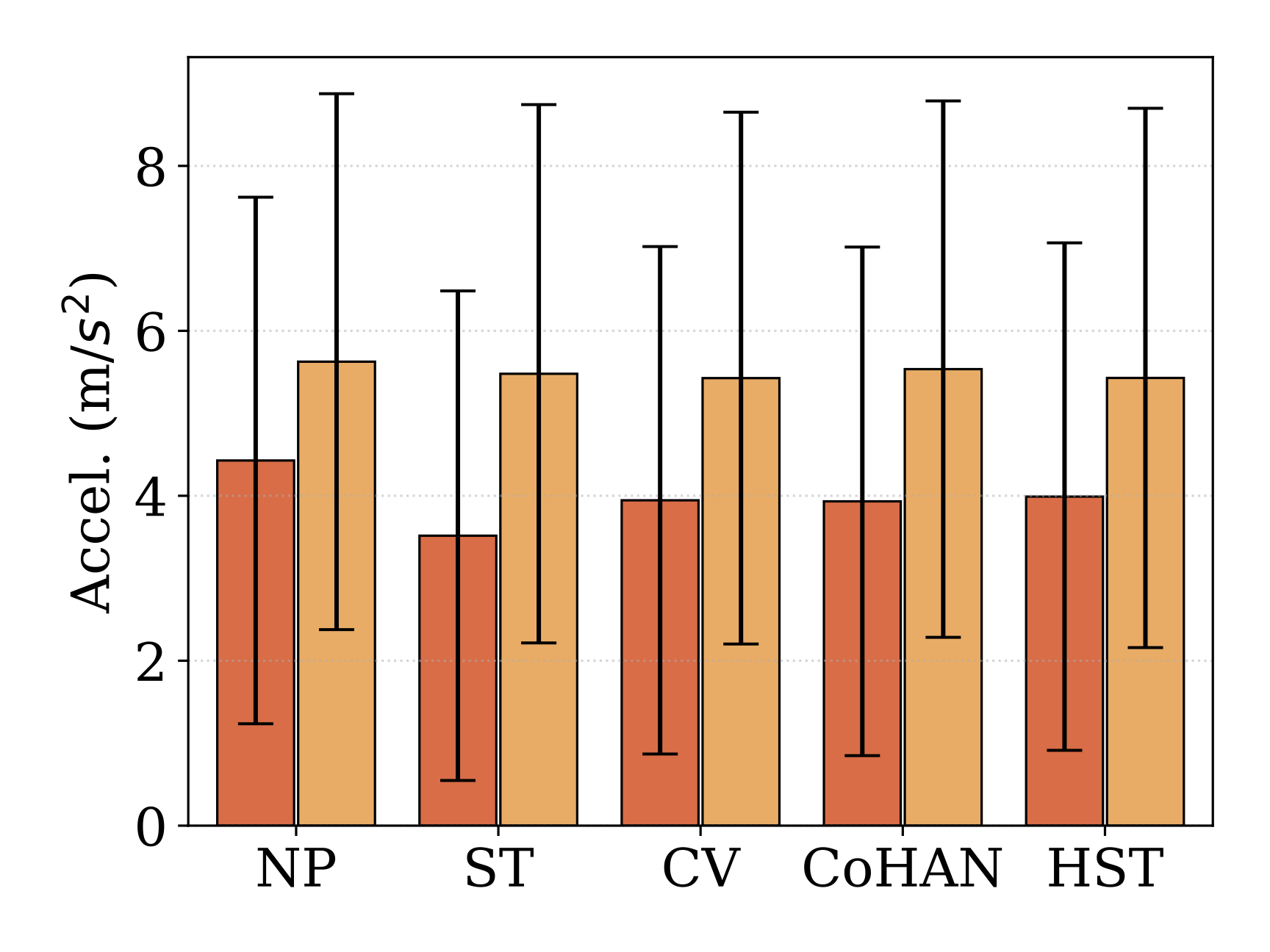}\\[6pt]
  \includegraphics[width=0.48\columnwidth]{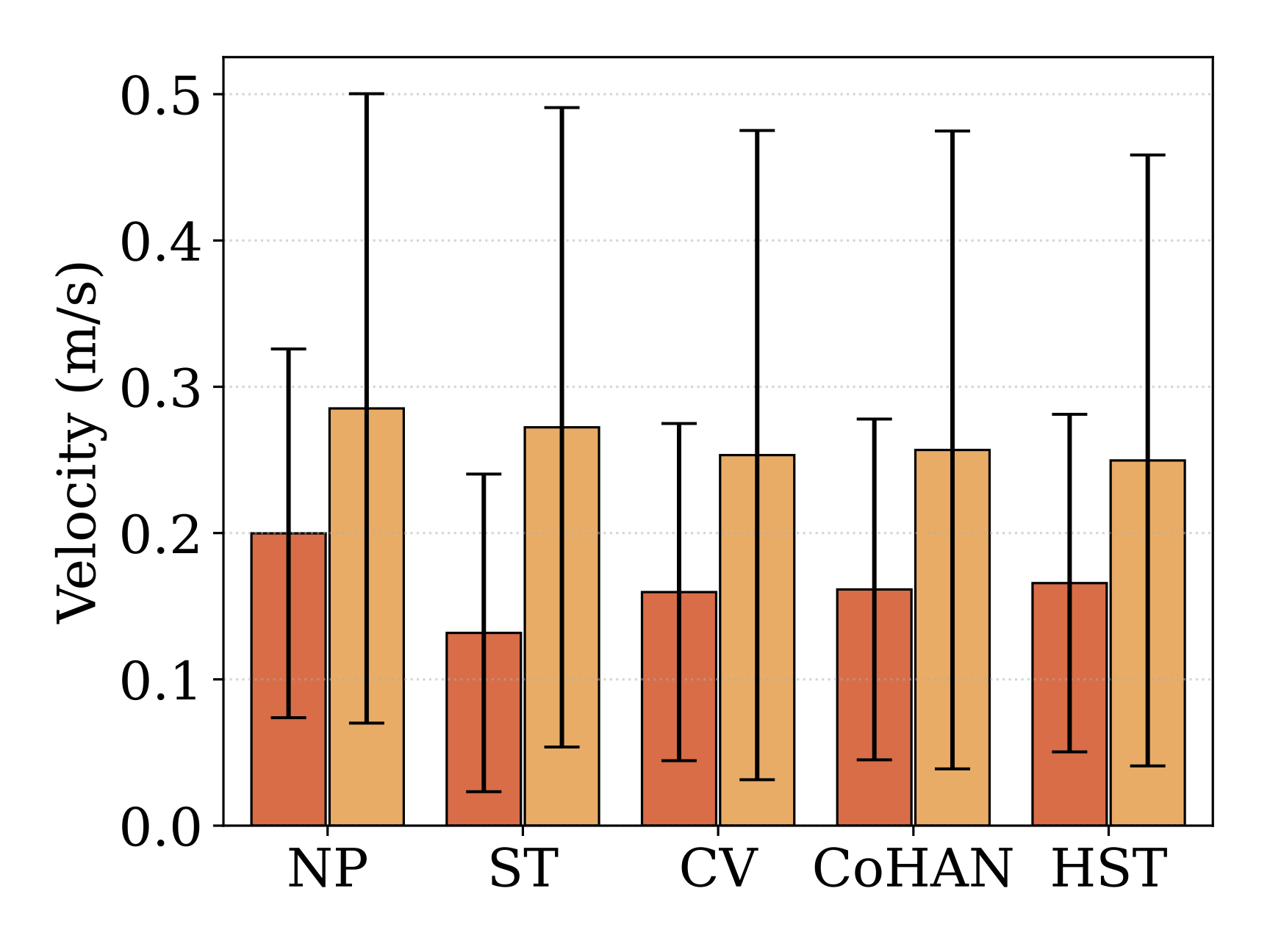}%
  \hfill
  \includegraphics[width=0.48\columnwidth]{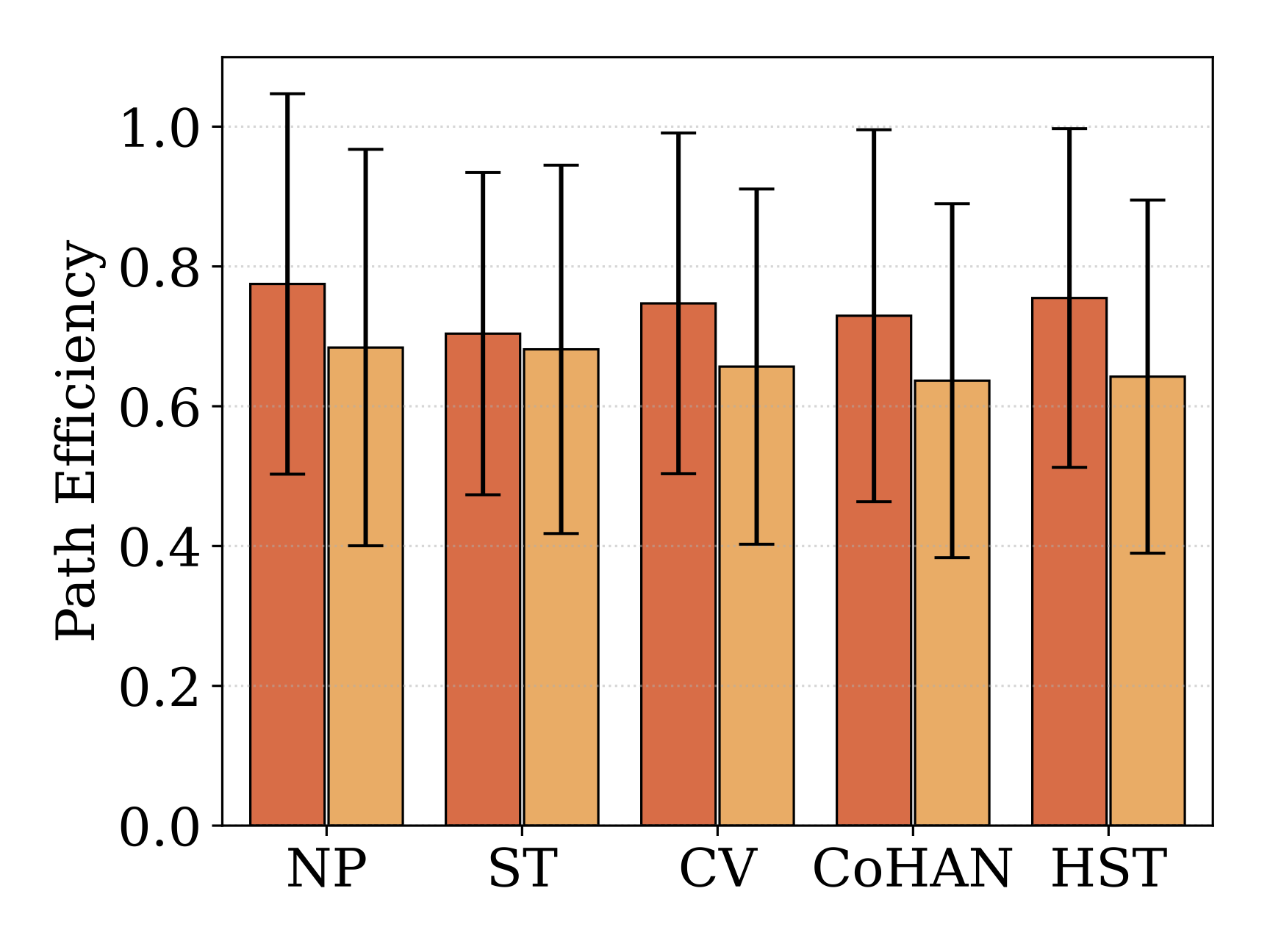}
  \caption{Robot motion statistics by controller and site (bars are means and errors are standard deviations). Dark orange refers to UM and light orange to LAAS.
  }
  \label{fig:robot_stats}
\end{figure}

Averages over the above robot metrics by algorithm and by site are in figure \ref{fig:robot_stats} (dataset names correspond to~\figref{tab:literature}). We identify that the robot had consistently higher distances from humans, faster and less smooth movement, and less direct paths at LAAS. Our control over the experimental setup leads us to speculate that the larger platform at LAAS caused the robot to need longer paths to avoid humans, or that cultural differences between the two sites may have played a role.

Within each site we see more limited diversity between controllers. In particular, we find that NP and ST have distinctly higher and lower values across all metrics than CV, CoHAN, and HST. Despite the differences in motion metrics for CV, CoHAN, and HST being more subtle, we find that they received vastly different user impressions. This highlights the importance of user impressions for identifying desirable SRN algorithm behavior.

\textbf{User Impressions and Comments}. We see that each robot behavior elicited vastly different impressions from users. The RoSAS and NASA TLX survey responses can be found in tables \ref{tab:rosas} and \ref{tab:tlx} respectively. We see substantial differences in mean both between algorithms and between sites, indicating each controller was received very differently by users. Furthermore, the consistently high standard deviations by trial serve to indicate that impressions varied strongly between each individual trial, a fact which shows the diversity of impressions exhibited in our dataset.

User comments also provide valuable descriptions of specific behavior that informed users' impressions of the robot. High level themes are substantially different by model. NP and CoHAN received the most negative comments, with many concerning them being \textit{``less responsive"} and \textit{``too close."} ST and CV received more balanced comments, however they still received comments describing them as \textit{``hard to predict"} (ST) and that \textit{``the robot ended up staying in annoying places"} (CV). HST received the most positive and least negative comments, with specific comments describing it as \textit{``very interactive"} and that it \textit{``was a lot smoother compared to the other ones."} Notably, all models received at least some positive and negative comments, which can help identify scenarios in which models that performed worse overall still achieved desirable behavior.

\begin{table}
\centering
\caption{RoSAS scale responses by algorithm (mean and standard deviation). The top row for each attribute refers to UM and the bottom row to LAAS. \textbf{Discomfort} is a mean of the first six rows, whereas \textbf{Competence} is a mean of the bottom six.}
\resizebox{\columnwidth}{!}{%
\renewcommand{\arraystretch}{0.8}
\begin{tabular}{llllllll}
  \toprule
Attribute & NP & ST & CV & CoHAN & HST \\ 
  \midrule
Aggressive &2.53$\pm$2.15&2.06$\pm$1.43&2.59$\pm$1.84&2.88$\pm$2.42&2.59$\pm$2.15\\
 &2.80$\pm$2.20&1.85$\pm$1.46&1.90$\pm$1.43&3.22$\pm$2.42&2.05$\pm$1.57\\
 \midrule
Awful 
&1.71$\pm$1.40&1.82$\pm$1.51&1.47$\pm$0.80&1.82$\pm$1.33&1.71$\pm$1.05\\
 &2.72$\pm$1.89&2.48$\pm$1.99&2.22$\pm$1.78&2.35$\pm$1.79&1.92$\pm$1.47\\
 \midrule
Awkward
&4.65$\pm$1.87&5.06$\pm$2.33&4.76$\pm$2.33&4.88$\pm$2.09&5.12$\pm$2.50\\
 &4.00$\pm$2.12&2.82$\pm$2.11&3.05$\pm$2.10&3.72$\pm$2.17&3.25$\pm$2.06\\
 \midrule
Dangerous &2.06$\pm$1.68&1.65$\pm$1.00&2.00$\pm$1.66&2.24$\pm$1.82&2.35$\pm$1.97\\
 &2.48$\pm$1.99&2.00$\pm$1.66&1.82$\pm$1.26&3.25$\pm$2.17&1.85$\pm$1.33\\
 \midrule
Scary
&1.82$\pm$1.55&1.71$\pm$1.10&1.82$\pm$1.42&1.94$\pm$1.71&1.88$\pm$1.62\\
 &2.72$\pm$2.04&2.02$\pm$1.66&1.88$\pm$1.18&3.00$\pm$2.10&1.98$\pm$1.46\\
 \midrule
Strange
&4.00$\pm$2.29&4.06$\pm$2.68&4.59$\pm$2.37&3.94$\pm$2.38&4.35$\pm$2.42\\
 &4.03$\pm$2.24&3.32$\pm$2.23&3.20$\pm$1.98&3.65$\pm$2.05&2.60$\pm$1.84\\
 \midrule
Knowledgeable &4.82$\pm$1.51&4.65$\pm$1.80&5.12$\pm$1.58&4.88$\pm$1.69&4.53$\pm$1.74\\
 &4.78$\pm$1.86&4.82$\pm$1.71&4.90$\pm$1.74&4.68$\pm$1.87&5.05$\pm$2.02\\
 \midrule
Interactive
&4.24$\pm$2.08&4.76$\pm$2.02&4.24$\pm$1.75&4.24$\pm$2.17&3.94$\pm$1.78\\
 &4.30$\pm$2.11&4.58$\pm$2.32&4.53$\pm$2.20&3.62$\pm$1.93&4.78$\pm$2.41\\
 \midrule
Responsive
&5.12$\pm$1.87&6.18$\pm$2.10&4.59$\pm$1.62&4.94$\pm$2.49&4.71$\pm$2.47\\
 &5.03$\pm$2.04&5.70$\pm$1.76&5.52$\pm$2.10&4.05$\pm$1.62&6.12$\pm$2.02\\
 \midrule
Capable &5.53$\pm$1.77&5.47$\pm$1.42&5.47$\pm$1.23&5.00$\pm$1.80&5.00$\pm$1.87\\
 &4.95$\pm$1.66&5.55$\pm$1.65&5.55$\pm$2.09&5.00$\pm$1.80&5.58$\pm$1.63\\
 \midrule
Reliable &4.94$\pm$1.68&5.06$\pm$1.34&5.12$\pm$1.45&4.76$\pm$1.99&4.65$\pm$1.97\\
 &5.20$\pm$1.87&5.38$\pm$1.78&5.60$\pm$1.91&4.95$\pm$1.83&5.38$\pm$1.97\\
 \midrule
Competent &5.76$\pm$1.44&4.88$\pm$1.73&4.82$\pm$1.63&4.76$\pm$1.56&4.53$\pm$1.77\\
 &4.80$\pm$1.68&5.28$\pm$1.88&5.52$\pm$2.00&4.60$\pm$1.81&5.28$\pm$2.04\\
 \midrule
\textbf{Discomfort} &2.79$\pm$1.07&2.73$\pm$1.06&2.87$\pm$1.14&2.95$\pm$1.50&3.00$\pm$1.38\\
 &3.12$\pm$1.60&2.42$\pm$1.49&2.35$\pm$1.05&3.20$\pm$1.56&2.28$\pm$1.18\\
 \midrule
\textbf{Competence} &5.07$\pm$1.32&5.17$\pm$1.20&4.89$\pm$0.99&4.76$\pm$1.50&4.56$\pm$1.70\\
 &4.84$\pm$1.53&5.22$\pm$1.36&5.27$\pm$1.59&4.48$\pm$1.37&5.36$\pm$1.46\\
\bottomrule
\end{tabular}
}
\label{tab:rosas}
\end{table}

\section{Anticipated Use Cases}

\textbf{Human motion prediction in close-quarters human-robot interactions:} HMP in close-quarters, task-driven settings is particularly challenging due to the highly irregular paths caused by forced close interactions, high velocities caused by the need for productivity, and rapid direction switches caused by moving between specific goals. Deployment in warehouses, hospitals, and airports makes high performance under such settings necessary. As demonstrated in Sec.~\ref{analysis}, our dataset captures these challenges, and thus presents a quality setting for developing such HMP algorithms. Furthermore, the human and robot goal information we include can aid in developing goal conditioned prediction models~\citep{mangalam_ynet} which may prove better in these settings. 

\begin{table}[t]
\centering
\caption{NASA TLX scale responses by algorithm (mean and standard deviation). The top row for each attribute are statistics from UM, while the bottom row are from LAAS.}

\resizebox{\columnwidth}{!}{%
\renewcommand{\arraystretch}{0.8}
\begin{tabular}{llllllll}
  \toprule
Workload Type & NP & ST & CV & CoHAN & HST \\ 
  \midrule
Mental
&2.24$\pm$2.91&2.12$\pm$2.57&2.35$\pm$3.12&2.71$\pm$3.04&2.82$\pm$3.38\\
 &2.45$\pm$2.59&2.65$\pm$2.57&2.40$\pm$2.58&3.30$\pm$3.25&2.38$\pm$2.70\\
 \midrule
Physical 
&2.82$\pm$3.28&3.00$\pm$2.76&3.18$\pm$3.73&2.94$\pm$2.66&2.71$\pm$2.66\\
 &2.20$\pm$2.37&1.92$\pm$2.19&2.22$\pm$2.37&2.50$\pm$2.59&2.08$\pm$2.10\\
 \midrule
Temporal 
&3.12$\pm$2.64&3.94$\pm$3.80&4.41$\pm$3.64&3.18$\pm$2.30&4.41$\pm$5.03\\
 &3.52$\pm$3.43&2.80$\pm$3.60&3.15$\pm$3.42&3.52$\pm$3.45&2.80$\pm$3.22\\
 \midrule
Performance &18.6$\pm$5.50&19.8$\pm$1.98&18.2$\pm$5.32&19.5$\pm$3.28&17.9$\pm$5.75\\
 &18.8$\pm$5.08&19.2$\pm$3.78&18.2$\pm$5.37&18.8$\pm$4.18&18.9$\pm$4.61\\
 \midrule
Effort 
&2.53$\pm$2.62&3.35$\pm$3.66&2.59$\pm$2.94&3.00$\pm$3.34&2.94$\pm$3.54\\
 &3.32$\pm$4.51&2.22$\pm$2.58&2.98$\pm$3.17&2.90$\pm$2.73&2.20$\pm$2.30\\
 \midrule
Frustration &1.94$\pm$2.90&2.47$\pm$3.41&2.35$\pm$2.40&3.24$\pm$4.37&3.65$\pm$5.16\\
 &3.90$\pm$4.33&2.30$\pm$2.88&3.15$\pm$3.98&4.25$\pm$4.64&2.38$\pm$3.10\\
\bottomrule
\end{tabular}
}

\label{tab:tlx}
\end{table}

\textbf{Learning robust social navigation policies}. The inclusion of a robot navigating among humans makes this dataset valuable for directly learning SRN policies or models which can be integrated into MPCs. In addition to containing desirable behavior, our dataset contains information on how humans react when the robot acts in error-- valuable data for learning policies that are robust to failure. Methods such as offline reinforcement learning \citep{Fujimoto2018OffPolicyDR} can be used to learn SRN policies that emulate desirable behavior while simultaneously learning to avoid failure modes in the data.

\textbf{Learning from human preferences}. Existing methods for preference-based learning in SRN have crowdsourced annotations of preferences over trajectory choices online \citep{wang2022feedbackefficient}. The lack of physical interaction between labelers and the robot introduces a fundamental discrepancy between the labelers' preferences and those who interact with the robot during deployment. The numerical survey responses in our dataset overcome this by providing impression labels directly from the users in the trajectory data immediately following their real-world SRN interactions. Techniques such as preference-based inverse reinforcement learning~\citep{christiano_rlhf} can be used with our data to learn reward functions matching the preferences of users who interact with the physical system.

\textbf{Identifying embodiment and cultural effects in social navigation}. It is inevitable that SRN systems will be deployed on different robots in areas with distinct cultural norms. Our dataset contains user interactions with both a lightweight, small robot and a heavy, large robot in two culturally distinct locations. Although both morphology and culture varying across sites limits our ability to attribute the observed differences to either specifically, the variation over these factors increases the applicability of our data to a broader range of future SRN deployment settings.

\footnotesize
\balance
\bibliography{references}
\bibliographystyle{abbrvnat}


\end{document}